\def\year{2022}\relax
\documentclass{article} 
\usepackage{aaai22}  
\usepackage{times}  
\usepackage{helvet}  
\usepackage{courier}  
\usepackage[hyphens]{url}  
\usepackage{graphicx} 
\usepackage{comment}
\usepackage{balance}
\usepackage{natbib}  
\usepackage{caption} 
\DeclareCaptionStyle{ruled}{labelfont=normalfont,labelsep=colon,strut=off} 
\usepackage{algorithm}
\usepackage{algorithmic}

\usepackage{bbm}
\usepackage{tikz}
\usepackage{amsmath} 
\usepackage{amssymb}
\usepackage{balance}
\usepackage{capt-of}

\usepackage{newfloat}
\usepackage{listings}
\floatstyle{ruled}
\newfloat{listing}{tb}{lst}{}
\floatname{listing}{Listing}
\title{Spatiotemporal Transformer for Stock Movement Prediction}
\author{
Daniel Boyle and Jugal Kalita\\
University of Colorado, Colorado Springs\\
1420 Austin Bluffs Pkwy\\
Colorado Springs, CO 80919\\
}

\usepackage{bibentry}

\begin{document}

\maketitle

\begin{abstract}
Financial markets are an intriguing place that offer investors the potential to gain large profits if timed correctly. Unfortunately, the dynamic, non-linear nature of financial markets makes it extremely hard to predict future price movements. Within the US stock exchange, there are a countless number of factors that play a role in the price of a company's stock, including but not limited to financial statements, social and news sentiment, overall market sentiment, political happenings and trading psychology. Correlating these factors is virtually impossible for a human. Therefore, we propose STST, a novel approach using a Spatiotemporal Transformer-LSTM model for stock movement prediction. Our model obtains accuracies of 63.707 and 56.879 percent against the ACL18 and KDD17 datasets, respectively. In addition, our model was used in simulation to determine its real-life applicability. It obtained a minimum of 10.41\% higher profit than the S\&P500 stock index, with a minimum annualized return of 31.24\%. 
\end{abstract}

\section{Introduction}

The stock market is a public financial market that allows companies the ability to sell stakes in their company in order to raise necessary capital from investors. From the investor's standpoint, it provides a place to buy and sell shares of a company or a collection of companies while providing them with partial ownership relative to the number of shares that are held. Excluding external factors, the prices of these shares fluctuate due to the nature of supply and demand. Investors typically buy stock when they feel that the price will go up, allowing them to gain profits from their investments. Subsequently, investors will sell their stock when they believe that the price will go down, preventing them from incurring future losses. 

This creates a very intriguing environment where investors can potentially gain large profits. Every day, billions of dollars are exchanged on each of the U.S. stock exchanges \cite{dunne2015stock}. Behind each transaction, investors hope to adequately predict price fluctuations in order to generate profits. Unfortunately, predicting when a stock will go up or down is an extremely complex task due to the dynamic, non-linear nature of the stock market. Outside of company fundamentals and the price movements of a stock, there are a number of factors that make prediction difficult, including public opinion, political situations, trading psychology, media sentiment, and even the current day of the week \cite{maberly1995eureka}.

Due to the complexity of trading and the potential profitability, there has been a large amount of research put towards the ability to predict and forecast price movements and trends. Traditionally, there are two main methods of analysis used for financial market prediction: technical analysis and fundamental analysis. The idea behind technical analysis is that price and volume data completely contain all the information needed to adequately predict the market. Therefore, it is focused on studying price movements and chart patterns to help detect trends, reversal patterns, and other types of technical signals \cite{edwards2018Technical}. Fundamental analysis on the other hand, is focused on measuring the intrinsic values of securities by analyzing company financial statements and various micro and macroeconomic factors \cite{abarbanell1997Fundamental}. Investors often use this type of analysis to look for securities that are mispriced in order to determine areas of investment. While there are many investors and experts that are strong proponents of each form of analysis, there also exist many sub-methodologies that combine both forms of analysis in order to give a broader look into the behavior of the stock market.

In recent years, significant advances in the field of machine learning for data processing and analysis have yielded encouraging outcomes when applied to complicated prediction and regression tasks. This has caught the eye of financial experts and analysts for use within financial markets. Recent research has applied machine learning methods such as Support Vector Machines (SVM) \cite{hearstSVM1998}, Boosted Trees \cite{chen2016xgboost}, and complex neural network models to different tasks within the financial market realm. These methods have become very popular due to their innate ability to process large amounts of data and extract complex features that would otherwise be impossible to obtain. Major focuses for the application of machine learning include trading signal prediction \cite{wu2020RlGruTrading}, price forecasting \cite{niaki2013ForecastingStocksAnn}, \cite{sen2017robust} and price movement prediction \cite{li2022Slat}.

With trading signal prediction, the goal is to predict the time or price at which to buy and sell a security in order to maximize profit and minimize loss. These points are commonly referred to as ``signals'' For price forecasting, the goal is to predict the future closing price or moving average of a security for a limited number of timesteps into the future. Lastly, price movement prediction is focused on predicting the direction of price movements across a specified time frame. 

For day traders, the objective is most commonly to try and accurately predict when the price of a stock might move up or down so that they can profit from price movements. Due to the number of external factors that can affect the stock market and the amount of data that is available at any given time for a specific security, it is incredibly difficult for a day trader to identify all the necessary patterns within the data in order to accurately predict future price movements. Due to their ability to learn complex long-range temporal relationships among data via self-attention, transformer models have been extremely popular in recent years across a wide variety of sequence-related problems. Using this notion, we propose the Spatiotemporal Transformer for Stocks (STST), a transformer-LSTM model that is focused on correlating spatiotemporal patterns for the task of stock movement prediction.

\section{Background}

Traditionally, Long Short-Term Memory Networks (LSTMs) \cite{hochreiter1997LSTM} and Convolutional Neural Networks (CNNs) \cite{hochreiter1997LSTM} have been the most commonly used models for sequence-based tasks. While they are both still very effective, they also have their downfalls. For LSTMs, data is fed in a sequential manner across timestep boundaries, which causes the training process to be extremely slow. This becomes a large issue when dealing with a significant amount of data. CNNs on the other hand, take in a sequence all at once, so they are significantly faster to train. Unfortunately, since CNNs ingest the entire sequence at once, they do not maintain a notion of position. Furthermore, they can struggle to efficiently learn long-term dependencies \cite {linsley2018longRangeDeps}. This can be overcome by the addition of larger receptive fields, but this can quickly become very computationally expensive. 

\subsection{Transformers}

To overcome many of these challenges, the Transformer neural network architecture was proposed by Vaswani et al. in 2017 at Google \cite{vaswani2017attention} for language-based sequence tasks. Their approach showed that a sequential model using self-attention could obtain state-of-the-art performance against sequence-to-sequence (seq2seq) tasks such as language translation, a task previously dominated by LSTMs. The original transformer architecture can be seen below in Figure \ref{fig:transformer}. 

 \begin{figure} [!ht]
  \centering
  \hspace*{-0.5cm}
  \includegraphics[trim={0 1.5cm 0 1.5cm},clip,scale=0.7]{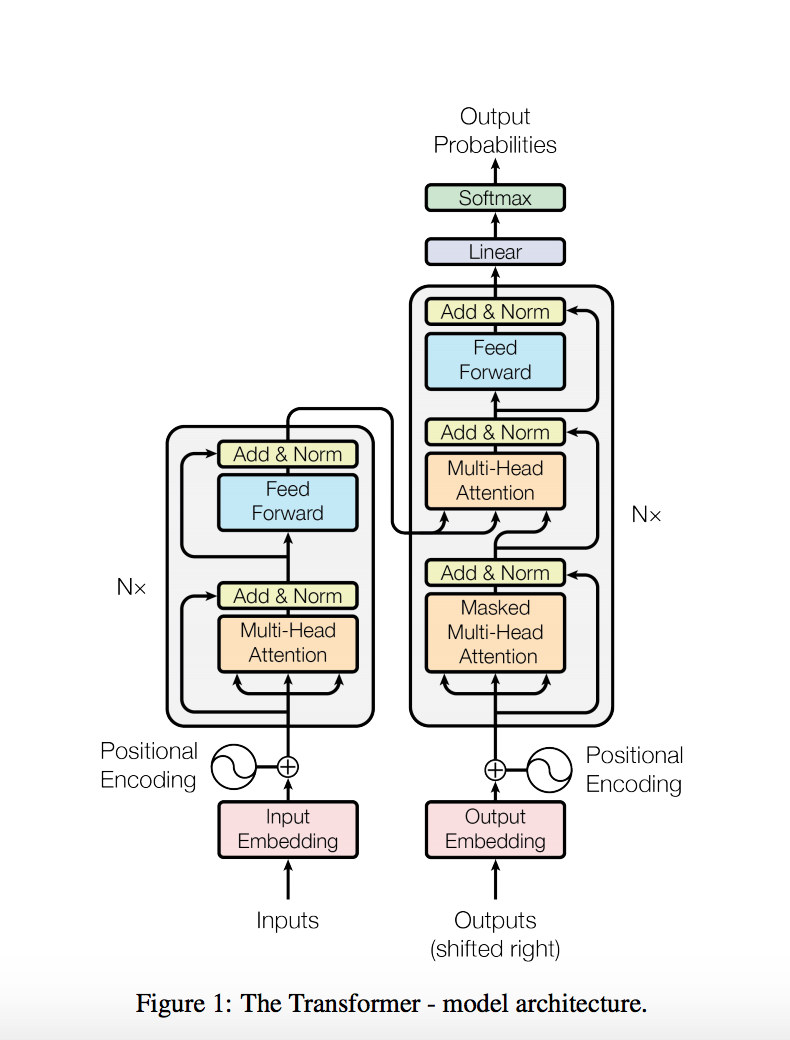}
  \caption{Original Transformer Architecture \cite{vaswani2017attention}}
  \label{fig:transformer}
\end{figure}

Similar to previous RNN-based architectures used for seq2seq tasks, the transformer is comprised of an encoder and a decoder, each containing stackable layers of attention and feed-forward neural networks. Within a transformer, there are five critical components: positional encoding, multi-headed attention, a feed-forward network, residual connections and normalization. 

\subsubsection{Positional Encoding}

In traditional seq2seq models, RNNs would inject the input one element at a time, forming a positional notion of the tokens in a sequence. For transformers, the entire input is fed into the network all at once; therefore, something needs to be added to the input feature vectors in order to encode a notion of position. This task is performed by a layer known as the positional encoder. In the original transformer, a fixed sinusoidal function was used to create a positional encoding that is summed with the original embedded input. In more recent years, positional embedding layers have gained popularity. They perform positional encoding using a trainable model. This can be seen used in well-known transformer-based language models such as RoBERTa \cite{liu2019roberta} and GPT-2 \cite{radford2019GPT2}.

\subsubsection{Multi-Head Self-Attention}

Arguably the most important component of the transformer architecture is multi-headed self-attention. Self-attention is a mechanism that is used to form a notion of importance between different variables within a sequence in order to help a model know what to pay attention to. Multi-headed self-attention allows for multiple different self-attention components (heads) that can learn distinct relationships within a sequence. This allows the transformer to learn complex long-term dependencies without many of the issues that plague Recurrent Neural Networks.

\subsubsection{Feed-Forward Network}

Similar to other neural network architectures, the transformer makes use of feed-forward, dense neural networks within the encoder and decoder positioned after the multi-headed attention. These layers map the output of the multi-head attention layer into the needed output vector space. This is often paired with a Rectified Linear Unit (ReLU) activation function \cite{nair2010relu} following the output of the network.

\subsubsection{Residual Connections}

Since transformers can often be extremely large, they are subject to the vanishing gradient problem. A common way to resolve this issue is to add residual connections within the network. A residual connection is essentially a way for data to skip over layers and be added or appended onto the output of a later layer \cite{he2016residual}. This helps the gradients propagate further back during back-propagation and allows for significantly larger networks.

\subsubsection{Normalization}

The last important layer of the transformer architecture is normalization. Normalization layers are traditionally used to help stabilize the dynamics of neural networks. The original transformer model uses a type of normalization known as layer normalization, focused on normalizing layer inputs across each dimension within a neural network. It was chosen over batch normalization, a technique commonly seen with large CNNs, due to the high variance across dimensions traditionally seen with word embeddings in Natural Language Processing (NLP) tasks \cite{vaswani2017attention}.

\section{Related Work}

In the following sections, we introduce related work on the use of transformers for modeling complex relationships, as well as research that has utilized transformers for tasks involving time-series data. We also look at previous research in the field of stock movement prediction. 

\subsubsection{Transformer-based Encoding}

Encoder-based transformers have grown in popularity in recent years as a method for efficiently learning latent representations for many different types of data. One of the most famous transformer-based models that has shown great promise in the field of representation learning is a model known as BERT, or Bidirectional Encoder Representations for Transformers, that was introduced in 2019 by Google \cite{devlin2019BERT}. BERT is a transformer-based encoder approach for learning contextualized bidirectional embeddings for tokens and sentences. Their approach introduced two novel unsupervised pre-training techniques to adequately contextualize tokens or sentences. The first method is known as Masked Language Modelling (MLM). MLM is focused on pre-training the encoder by masking out random tokens or groupings of tokens within the input, and using the model to try to predict the missing tokens. Secondly, they introduced the Next Sentence Prediction (NSP) pre-training technique that focused on predicting whether or not a specific sentence follows another sentence. BERT was able to obtain state-of-the-art results for 11 different NLP-related tasks.

Moving towards the time-series realm, in 2021, Zerveas et al. proposed the use of a transformer-encoder-based model to perform unsupervised representation learning for multi-variate time series \cite{zerveas2021MultiVarRepLearning}. Their approach utilized a similar method to BERT's MLM pre-training approach by masking out random variables within a sequence and having the model try to predict them. Their approach was able to pass the current state-of-the-art models for both classification and regression across several different multi-variate time series datasets.

Furthermore, Dang et al. used a BERT-based approach for time-series named TS-BERT for performing anomaly detection for uni-variate time-series data \cite{dangTsBertAnomoly}. Unlike Zerveas et al., which used an embedding layer to encode the time-series, TS-BERT uses a time-series dictionary to create an aggregate sequence that can be converted and used with the traditional BERT model. TS-BERT was able to beat out the current state-of-the-art in time-series anomaly detection. 

\subsubsection{Stock Movement Prediction}

The ability to predict the direction of a stock's future movement, commonly referred to as ``stock movement prediction,'' has captivated both investors and researchers due to the complex nature of the problem and the large potential for monetary gain. Most methods for stock market prediction can be divided into two categories: those that use only price data and those that combine price data with other external information. This type of information can include items such as news articles or social media posts. For this paper, we are focused solely on related work that utilizes only historical price data.

In 2018, Feng et al. introduced Adv-ALSTM, a temporal attention-based Long Short Term Memory (LSTM) network \cite{feng2018advALSTM} based on Qin et al.'s ALSTM model \cite{qin2017ALSTM}. They used adversarial training through the use of added perturbations in order to simulate the stochasticity of price data in the stock market. Their approach was able to obtain higher accuracy and MCC scores than StockNet \cite{Xu2018StockNet}, the state-of-the-art model at the time. The model was evaluated against the ACL18 and KDD17 datasets, two commonly used datasets for the task of price movement prediction, where it obtained accuracies of 57.20\% and 53.05\%, respectively. 

Zhang et al. proposed an approach in 2021 called FA-CNN that combines a deep factorization model known as DeepFM \cite{guo2018DeepFm} with attention-based CNN (ATT-CNN). They utilize the DeepFM model to extract spatial intra-day features, while applying the ATT-CNN model for the extraction of temporal features across time points \cite{zhangFmCnn2021}. They applied their model against three datasets for the Food and Beverage, Health Care and Real-Estate divisions within the Chinese stock market. Their model obtained state-of-the-art results against each of the datasets. 

Following shortly after, Yoo et al. proposed a model known as the Data-axis Transformer with Multi-Level Contexts (DTML) \cite{yoo2021LstmTransformer} for stock movement prediction. Their approach feeds multiple individual stocks and a global market index through an LSTM with temporal attention to capture a context vector for each input. The context vectors for each stock and the global market index are then stacked together to provide a multi-level context for the current market state. The result is then fed through a transformer encoder to correlate the different contexts and produce an overall prediction. They tested their dataset against three foreign-based datasets and three US-based datasets. This includes the ACL18 and KDD17 datasets. They achieved an accuracy of 57.44\% and 53.52\%, respectively, beating the state-of-the-art at the time.

In early 2022, Li et al. proposed a selective transfer learning approach known as STLAT that uses an LSTM with temporal attention paired with adversarial training for stock movement prediction \cite{li2022Slat}. By selectively considering what portion of the source data is useful and pairing this with adversarial training, it allows them to further increase the accuracy of their approach beyond the base attentive LSTM model. Their model was tested against the ACL18 and KDD17 datasets, becoming the current state-of-the-art after achieving accuracies of 64.56\% and 58.27\%, respectively. 

Later, in mid-2022, He et al. proposed Instance-based Deep Transfer Learning with Attention (IDTLA) \cite{he2022InstanceBased}. Their approach selects the most relevant source data for training based on a larger dataset. This allows for models to be trained effectively against stocks that have less available price data. For comparison with state-of-the-art methods, they trained their model against sector-based subsets of the ACL18 dataset. They obtained an overall accuracy of 58.2\% and an MCC score of 0.168, calculated by averaging the results on each individual dataset. While not overcoming the state-of-the-art, they introduce a novel approach to training models with relatively high accuracy against stocks with minimal historical data.

\begin{figure*} [!htb]
  \centering
  \hspace*{0.1cm}
  \includegraphics[trim={0 0.5cm 0 0.2cm},clip,scale=0.27]{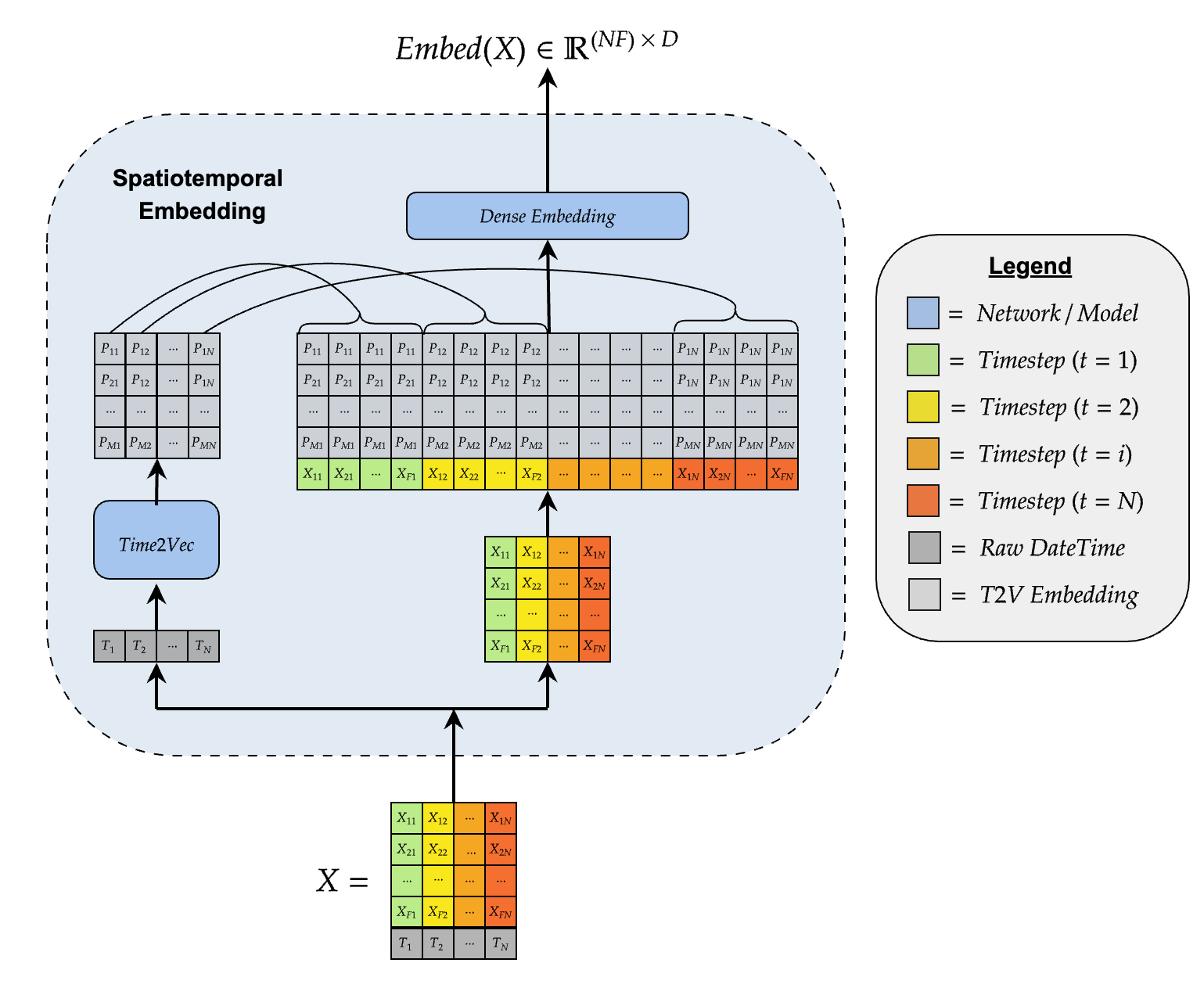}
  \caption{Spatiotemporal embedding} 
  \label{fig:spatiotemporal}
\end{figure*}

\section{Problem Formulation}

The goal of stock movement prediction is to learn a function $f(\hat{x}, \theta, T) \rightarrow \{0, 1\}$, where $\hat{x}$ represents the input vector $\textless x_1, x_2, ... , x_t\textgreater$ with $x_i \in \mathbb{R}$, $\theta$ represents the model parameters, $T$ represents a configurable time horizon and $\{0, 1\}$ represents the price movement direction. Therefore, given an input $\hat{x}$ for a stock, the objective is to predict whether the stock price will go up ($1$) or down ($0$) after $T$ timesteps. The goal of our research, as well as the related work mentioned earlier, is focused on the use of the time horizon, $T=1$. That is, we are interested in predicting whether the price will increase or decrease in the next trading day.

\section{Methodology}


A large portion of algorithmic and technical trading involves identifying key characteristics and patterns among historical price data and technical indicators. These patterns may occur close to each other or may be separated by several weeks. Thus, it is important that our model is able to find and capture these patterns successfully. To do this, we chose to use a transformer due to their ability to capture complex long-term dependencies through the use of self-attention. In the following sections, we introduce the Spatiotemporal Stock Transformer (STST), a Transformer-LSTM model for performing stock movement prediction.

\subsection{Model}

\begin{figure*} [!htb]
  \centering
  \hspace*{0.1cm}
  \includegraphics[trim={0 0.5cm 0 0.2cm},clip,scale=0.27]{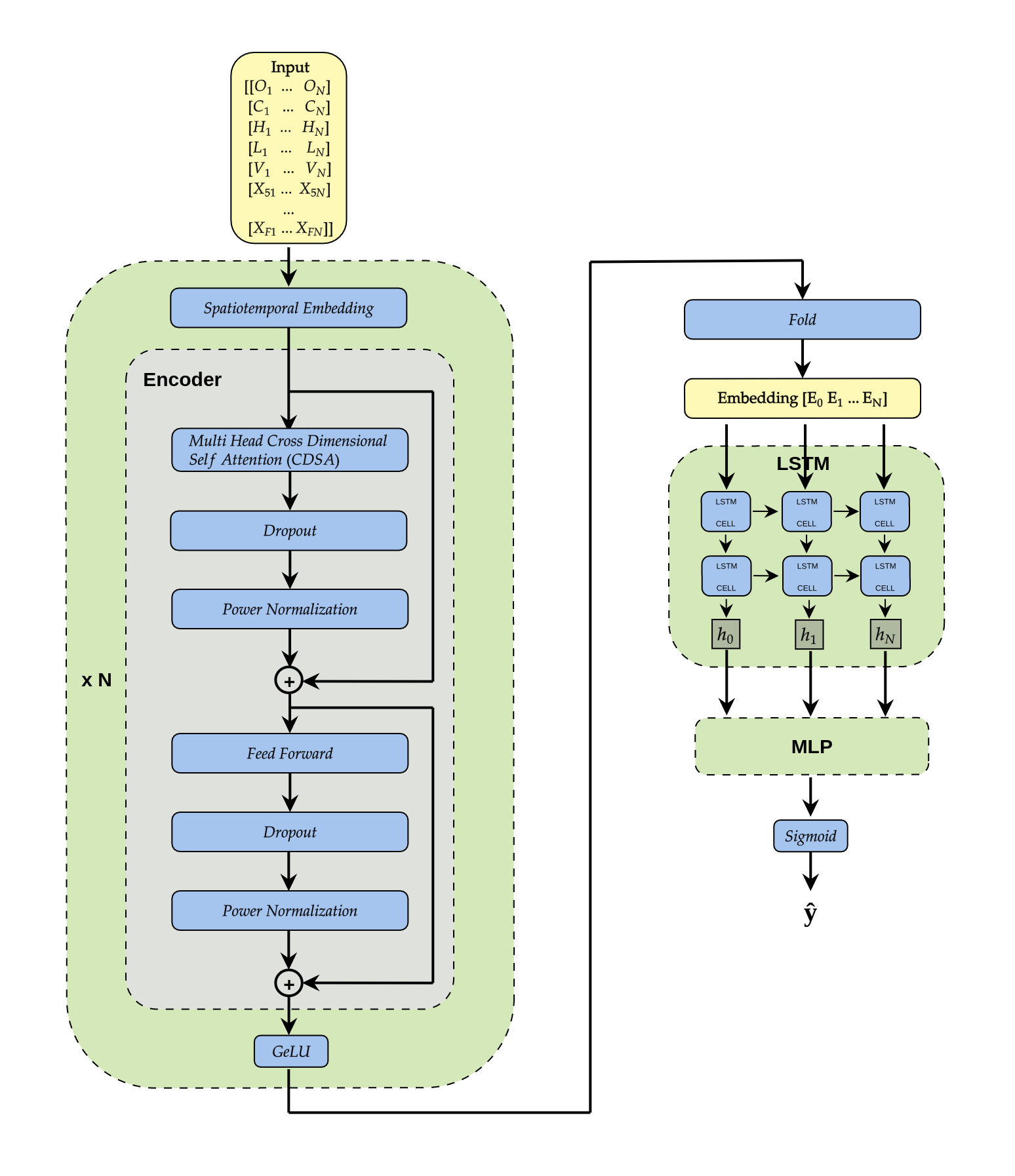}
  \caption{STST: Spatiotemporal Stock Transformer}
  \label{fig:model}
\end{figure*}

Our model utilizes a transformer-based encoder that is used to capture contextualized embeddings for individual stocks across a specified context window $W$. These contextual embeddings are fed into a multi-layered LSTM that is paired with a dense neural network for making the final predictions. On top of the traditional transformer, there are a few aspects that we have modified, including the positional encoding and embedding layers, in order to improve the performance of the model. The STST model can be seen in Figure \ref{fig:model}.

\subsubsection{Positional Encoding}

Instead of using traditional sinusoidal positional encoding or the more recent trainable positional encoders, we chose to use Date2Vec, a pre-trained variant of a time-based encoder called Time2Vec that has been shown to improve results in time-series-based prediction tasks \cite{kazemi2019Time2Vec}. Similar to sinusoidal and learned positional encoding layers, Time2Vec aims to create orthogonal time-embeddings. This makes it possible to add the positional vector to a feature vector without altering the original meaning of the input feature vector.

\subsubsection{Spatiotemporal Embedding and Self-Attention}

Each of the most recent state-of-the-art methods for stock prediction utilizes temporal attention. This can be very useful in that it helps the model learn to pay attention to certain time-based events in order to make a better prediction. However, temporal attention has the disadvantage of only focusing on how to connect relationships between different timesteps in a sequence. While this is useful, it does not take into account possible spatial correlations between different features or other relationships that may exist within the same timestep. For example, the difference between the opening, closing, high, and low prices within a single timestep can provide information that may be useful to day traders when making decisions. These are conditions that exist along spatial boundaries and are not easily captured by traditional temporal attention. 

In order to do this, we create a special spatiotemporal embedding layer that transforms the input features into a flattened feature vector. The details on this layer follow below. Let each input be represented by $X \in \mathbb{R}^{N \times (T+F)}$, where $N$ is the context window size, $F$ is the number of non-time-based features, and $T$ is the number of time-based features. The spatiotemporal embedding splits $X$ into $X_{f} \in \mathbb{R}^{N \times F}$ and $X_{t} \in \mathbb{R}^{N \times T}$. The positional encoding is then applied with $P_{t} = Time2Vec(X_{t})$. The feature matrix $X_f$ is then flattened into $X'_{f}$ such that $X'_{f} \in \mathbb{R}^{1 \times (NF)}$. Once flattened, it is concatenated with $X_t$ to give resultant $X_c$, where $X_c = \{ x^{'}_{fi} \oplus x_{ti} | x^{'}_{fi} \in X^{'}_f, x_{ti} \in X_t \}$. Once concatenated, the resultant vector is passed through a Dense Neural Network with an output size of $D$ to get $E_X = Dense(X_c)$, with $E_X \in \mathbb{R}^{NF \times D}$. A visual representation of this can be seen in Figure \ref{fig:spatiotemporal}. 

Once the input $X$ is transformed into $E_X$, it is passed to the transformer-encoder, which contains a traditional multi-headed self-attention module, commonly seen in the original transformer \cite{vaswani2017attention}. The resulting self-attention map has a dimension of ${NF}\times{NF}$. This allows the multi-headed attention module to be able to capture relationships among individual features across both spatial and temporal boundaries. For evaluation, we compare the use of this spatiotemporal embedding with a version of our model using traditional embedding to observe its impact on the model's classification results. 

\subsubsection{Encoder}

Similar to the original transformer, our encoder is made up of a configurable number of blocks, each containing multi-headed attention, normalization, a feed-forward neural network and residual connections. However, the original transformer uses layer normalization, which usually does not provide good performance for tasks involving time-series data. Instead, we chose to use power normalization, a reimagined form of batch normalization that overcomes many of the issues caused by batch normalization \cite{shen2020powernorm}. During our model-parameter search, we test to determine which type of normalization produces the best results for our problem. We found that power normalization consistently provided more stable results than batch or layer normalization. In addition to the type of normalization, we also investigate the best placement for normalization. The original transformer places normalization immediately after the self-attention and position-wise feed-forward networks. However, some research suggests that placing the normalization before these layers produces better results \cite{xiong2020PreLayerNorm}. Our model search found that traditional post-normalization best suits our purpose.

Following the series of transformer encoder blocks, the encoder output $Y_E$ has the shape $NF \times D$ due to the result of unrolling the spatial features in our embedding layer. However, we want the final output before the classification layer to contain a single context vector of length C for each timestep in the context window. Therefore, we re-stack the spatial features in $Y_E$ into a new output $Y_{E2}$ with shape $N \times C$ where $C \in \mathbb{R}^{FD}$.

\subsubsection{Classifier}

We compare and test two different variants of our model, each using a slightly different classification model on top of the transformer-encoder. The first model we test employs a two-layer feed-forward neural network. The second model that we test utilizes a multi-layered LSTM followed by a single feed-forward neural network. The LSTM has shown promising results for stock-market prediction using raw features, which were used in each of the previous state-of-the-art approaches \cite{feng2018advALSTM} \cite{yoo2021LstmTransformer} \cite{li2022Slat}. However, instead of feeding the raw features directly into the LSTM, we feed it with the encoded contextual features that are output from the transformer-encoder. Since price movement prediction is a binary classification problem, the output for each classification model is a single value. It is passed through the sigmoid activation function to obtain the output prediction $\hat{y} \in [0, 1].$ 

\subsubsection{Training}

\def\arraystretch{1.2}
\begin{table}[!htp]
\centering
\begin{tabular}{l|c}
\hline \hline
\textbf{Parameter} & \textbf{Search Space} \\ \hline \hline
learning\_rate                &  [1e-7 to 1e-4]       \\
batch\_size                   &  [16, 32, 64, 128]    \\
warmup\_steps                 &  [0:50000,5000]      \\
context\_window\_size          &  [4,8,16,32,64]       \\
n\_encoders                   &  [1,2,4,8]            \\
n\_heads                      &  [1,2,4,8]            \\
d\_model                     &   [32,64,128,256]       \\
d\_ff                        &   [256,512,1024, 2048]  \\
ff\_dropout                  &   [0 to 0.5]            \\
attn\_dropout                &   [0 to 0.5]            \\
n\_lstm\_layers              &   [1,2,4]               \\
d\_lstm\_hidden              &   [128,256,512]         \\
norm\_type                   &   [batch, layer, power] \\
norm\_placement              &   [pre, post]           \\ \hline \hline
\end{tabular}
\caption{Hyper-parameter selection for the STST model. \label{tbl:paramsearch}}
\end{table}

The STST model is trained for 100 epochs and utilizes early stopping to prevent the model from overfitting. To further improve training, we use a learning rate warm-up stage that linearly increases the learning rate from 0 to the maximum learning rate, $\lambda$ for a specified number of steps. The use of warm-up is a common tactic used to improve the stability transformers during training \cite{popel2018trainingTransformers}.

To tune the model, we perform a grid-based hyper-parameter search to determine the best combination of parameters. The search space for each hyper-parameter can be seen in Table \ref{tbl:paramsearch}. Due to the complexity of our model, there are a number of different hyper-parameters that need to be tuned to ensure the highest performance. The hyper-parameters can be broken down into three categories: training parameters, size parameters, and architectural parameters. The training hyper-parameters that we focus on are the learning rate, batch size, number of warm-up steps needed, and the dropout rates (attn\_dropout, ff\_dropout) for the attention and feed-forward networks. Next, we have the size parameters. These are focused on the size of different aspects of the model. We fine-tune a number of different size parameters, including the input context window size, the number of encoder blocks (n\_encoders), the number of attention heads (n\_heads), the encoded output feature vector size for the transformer (d\_model), the size of nodes in each dense layer (d\_ff), the number of LSTM layers (n\_lstm\_layers) and the hidden size for the LSTM network (d\_lstm\_hidden). The last category of parameters that we tune are the architectural parameters. The first architectural parameter is the type of normalization layer to use. For this, we test batch, layer and power normalization. The last architectural parameter is normalization placement. We determine whether the model performs best with normalization placed before (pre) or after (post) the self-attention and position-wise feed-forward networks inside the transformer encoder.

\section{Data}

\def\arraystretch{1.6}
\begin{table*}[]
\begin{tabular}{lll}
\hline \hline
\textbf{Feature}           & \textbf{Description}                                                                                                                                             & \textbf{Formula}                                                                                                                                                                    \\ \hline \hline
Year                       & Decimal value representing the Year                                                                                                                              & curr\_year / 3000                                                                                                                                                                   \\ \hline
Month                      & Decimal value representing the Month                                                                                                                             & curr\_month / 12                                                                                                                                                                    \\ \hline
Day                        & Decimal value representing the Day                                                                                                                               & curr\_day / 31                                                                                                                                                                      \\ \hline
Weekday                    & Decimal value representing the Day of the week                                                                                                                   & curr\_weekday / 7                                                                                                                                                                   \\ \hline
Open                       & Opening value of the day                                                                                                                                         & -                                                                                                                                                                                   \\ \hline
Close                      & Closing value of the day                                                                                                                                         & -                                                                                                                                                                                   \\ \hline
Adjusted Close             & Closing value of the day,  adjusted for stock splits                                                                                                             & -                                                                                                                                                                                   \\ \hline
High                       & Highest value of the day                                                                                                                                         & -                                                                                                                                                                                   \\ \hline
Low                        & Lowest value of the day                                                                                                                                          & -                                                                                                                                                                                   \\ \hline
Volume                     & Volume for the day                                                                                                                                               & -                                                                                                                                                                                   \\ \hline \hline
SIG\_SMA (10, 30, 50, 200) & \begin{tabular}[c]{@{}l@{}}Binary signal based on N-day simple moving average \\ (SMA) for N = 10, 30, 50 and 200.\end{tabular}                                   & \begin{tabular}[c]{@{}l@{}} $\begin{cases} 1, \;\;\;Close >  SMA \\ 0, \;\;\;otherwise \end{cases}$ \end{tabular}                                                                                               \\ \hline
SIG\_EMA (10, 30, 50, 200) & \begin{tabular}[c]{@{}l@{}}Binary signal based on N-day exponential moving \\ average (EMA) for N = 10, 30, 50 and 200.\end{tabular}                             & \begin{tabular}[c]{@{}l@{}}$\begin{cases} 1, \;\;\;Close >  EMA \\ 0, \;\;\;otherwise \end{cases}$ \end{tabular}                                                                                              \\ \hline
SIG\_MOMENTUM              & Binary signal based on 10-day Momentum indicator                                                                                                                 & \begin{tabular}[c]{@{}l@{}}$\begin{cases} 1, \;\;\;MOM > 0 \\ 0, \;\;\;otherwise \end{cases}$\end{tabular}                                                                                                    \\ \hline
SIG\_STOCHRSI              & \begin{tabular}[c]{@{}l@{}}Binary signal based on Stochastic Relative Strength \\ Index (STOCHRSI) (len=14, rsi\_len=14, k=3, d=3)\end{tabular}            & \begin{tabular}[c]{@{}l@{}} $\begin{cases} 1, \;\;\;SRSI(i) \le 25 \: or \\ \;\;\;\;\;\; SRSI(i) > SRSI(i-1) \: \\ \;\;\;\;\;\; and \: SRSI < 75 \\ 0, \;\;\;otherwise \end{cases}$ \end{tabular} \\ \hline 
SIG\_STOCH\_D              & Binary signal based on Stochastic D band                                                                                                                         & \begin{tabular}[c]{@{}l@{}}$\begin{cases} 1, \;\;\;STOD(i) > STOD(i-1) \\ 0, \;\;\;otherwise \end{cases}$\end{tabular}                                                                           \\ \hline
SIG\_STOCH\_K              & Binary signal based on Stochastic K band                                                                                                                         & \begin{tabular}[c]{@{}l@{}}$\begin{cases} 1, \;\;\;STOK(i) > STOK(i-1) \\ 0, \;\;\;otherwise \end{cases}$\end{tabular}                                                                          \\ \hline
SIG\_MACD                  & \begin{tabular}[c]{@{}l@{}}Binary signal based on Moving Average Convergence \\ Divergence (MACD) indicator \\ (fast=12, slow=26, signal=9)\end{tabular} & \begin{tabular}[c]{@{}l@{}}$\begin{cases} 1, \;\;\; MACD_{signal} < MACD \\ 0,\;\;\; otherwise \end{cases}$\end{tabular}                                                                                           \\ \hline
SIG\_CCI                   & \begin{tabular}[c]{@{}l@{}}Binary signal based on Commodity Channel Index \\ (CCI) indicator (14-day)\end{tabular}                                               & \begin{tabular}[c]{@{}l@{}}$\begin{cases} 1, \;\;\;CCI(i) \leq 100 \: or \\ \;\;\;\;\;\; CCI(i) > CCI(i-1) \\ 0, \;\;\;otherwise \end{cases}$\end{tabular}                                                  \\ \hline
SIG\_MFI                   & \begin{tabular}[c]{@{}l@{}}Binary signal based on 14-day Money Flow Index \\ (MFI) indicator\end{tabular}                                                         & \begin{tabular}[c]{@{}l@{}}$\begin{cases} 1, \;\;\;MFI(i) \leq 20 \: or \\ \;\;\;\;\;\; MFI(i) > MFI(i-1) \: \\ \;\;\;\;\;\; and \: MFI(i) < 80 \\ 0, \;\;\;otherwise \end{cases}$\end{tabular}                         \\ \hline
SIG\_AD                    & \begin{tabular}[c]{@{}l@{}}Binary signal based on Accumulation/Distribution \\ (AD) Index indicator\end{tabular}                                                  & \begin{tabular}[c]{@{}l@{}}$\begin{cases} 1, \;\;\;AD(i) > AD(i)-1 \\ 0, \;\;\;otherwise \end{cases}$\end{tabular}                                                                                       \\ \hline
SIG\_OBV                   & \begin{tabular}[c]{@{}l@{}}Binary signal based on On Balance Volume (OBV) \\ indicator\end{tabular}                                                                                                             & \begin{tabular}[c]{@{}l@{}}$\begin{cases} 1, \;\;\;OBV(i) > OBV(i)-1 \\ 0, \;\;\;otherwise \end{cases}$\end{tabular}                                                                                    \\ \hline
SIG\_ROC                   & \begin{tabular}[c]{@{}l@{}}Binary signal based on 10-day Rate of Change (ROC) \\ indicator \end{tabular}                                                                                                     & \begin{tabular}[c]{@{}l@{}}$\begin{cases} 1, \;\;\;ROC(i) > ROC(i)-1 \\ 0, \;\;\;otherwise \end{cases}$\end{tabular}                                                                                  \\ \hline \hline
\end{tabular}
\caption{STST Technical Input Feature Definitions}
\label{tbl:features}
\end{table*}

\subsection{Datasets}

To test the performance of our model, we chose the ACL18 \cite{Xu2018StockNet} and KDD17 
\cite{zhang2017KDD17} datasets. These datasets were selected because they are publicly available and are commonly used among studies for comparing model performance with respect to price movement prediction for stocks. 

The ACL18 dataset contains data from 88 high-volume stocks obtained from the US NASDAQ and NYSE exchanges. It contains historical price data ranging from January 1, 2014, until January 1, 2016. For data splits, the ACL18 dataset is further broken down, with the training set ranging from January 1, 2014, to August 8, 2015; the validation set ranging from August 8, 2015, to October 1, 2015; and the test set ranging from October 1, 2015, to January 1, 2016. 

For KDD17, the dataset contains data from the 50 highest-volume stocks obtained from the NASDAQ and NYSE exchanges. The KDD dataset contains historical price data ranging from January 1, 2007 to January 1, 2017. For data splits, the KDD17 dataset is further broken down, with the training set ranging from January 1, 2007 to January 1, 2015; the validation set ranging from January 1, 2015 to January 1, 2016; and the test set ranging from January 1, 2016 to January 1, 2017.

Both datasets are comprised of historical price tick data. This includes the open, close, adjusted close, high, and low prices, along with the trading volume on a per-day basis between the dates provided, not including weekends. On top of the basic historical price data that is provided, we generate additional aggregated data based on the price data; these are commonly known as technical indicators. This includes everything from simple moving averages to more complicated calculations for momentum and volatility. For our approach, we generate 22 additional features. The first four are generated by splitting apart the date into day, month, year, and weekday. These features are fed into the time2vec model and used for encoding position within each timestep. For the remaining features, we use 18 calculated binary signals based on common technical indicators. The details of each feature included and their calculations can be seen highlighted in Table \ref{tbl:features}. 

For both datasets, the output labels $Y$ are calculated based on the following thresholds in Equation 1 below. All data points that fall within the -0.5\% to 0.55\% range are ignored. These thresholds are chosen to help balance the datasets. This leaves a split of 50.64\% and 49.36\% split between the two classes for the ACL18 dataset, and a split of 50.70\% and 49.30\% for the KDD17 dataset. 
\begin{equation}
    \begin{cases} 
        0, \;\;\; (AdjClose_{t+1} / AdjClose_{t} - 1) <= -0.5\% \\ 
        1, \;\;\; (AdjClose_{t+1} / AdjClose_{t} - 1) >= 0.55\% \\ 
    \end{cases} 
\end{equation}

Other types of information that can be useful for stock price prediction include sentiment data, foreign market data, and company financial statements. While it has been observed in other research that these additional items can help improve the results of these tasks \cite{sawhney2020mansf}, they are often hard to obtain in large quantities, which makes it significantly more difficult to create a generalized model. However, our focus in this paper is on the use of only historical data for predicting movement. Therefore, we focus only on the use of price-based data. 

In addition to the dimensionality and contents of the input feature vectors, another aspect of the data that is crucial for our task is the context window size. The context window represents the number of timesteps that the model is able to see at once. Too small, and a model can have a difficult time obtaining enough contextual information to make informed decisions. If the window is too large, it can make it harder for the model to learn. For our model, the context window size is a hyper-parameter that is fine-tuned to balance performance and model convergence.

\section{Evaluation}

To evaluate the efficacy of the STST model, we present three experimental questions to answer in our research:

\begin{itemize} \setlength{\itemindent}{2em}
    \item[\textbf{Q1.}] \textbf{Market Prediction}: How does STST compare to other state-of-the-art approaches for stock market price prediction?
    \item[\textbf{Q2.}] \textbf{Ablation Study}: How does STST perform without several key-aspects of the approach? 
    \item[\textbf{Q3.}] \textbf{Investment Simulation}: How does STST perform when utilized for making trading decisions in the stock market? 
\end{itemize}

We outline the details for each experiment in the following sections. 

\subsection{Q1: Model Performance}

For baseline comparison, we test our model against three different types of approaches. The first type is randomness. This is performed by comparing our model's performance against a random classifier. This gives a basis for showing that our model is able to perform better than random. Next, we test our model against a simple baseline model. Since LSTM networks are very commonly used for sequence-to-sequence tasks and other stock-related problems, we use a multi-layered LSTM as the baseline model. The LSTM is fine-tuned to obtain the highest possible accuracy against each target dataset. Finally, we compare our model with other common state-of-the-art methods for stock movement prediction. Below is a list of each model that we use for comparison:

\begin{itemize} \setlength{\itemindent}{1em}
    \item \textbf{Adv-LSTM} \cite{feng2018advALSTM}: Temporal Attention-based LSTM combined with Adversarial Training.
    \item \textbf{MAN-SF} \cite{sawhney2020mansf}: Combines chaotic temporal signals from price data, social media and stock correlations using a Graph Neural Network (GNN).
    \item \textbf{DTML} \cite{yoo2021LstmTransformer}: Utilizes attention-based LSTM for individual stock contextualization, followed by a transformer for correlating the stocks and overall market to obtain final predictions.  
    \item \textbf{IDTLA} \cite{he2022InstanceBased}: Instance-based deep transfer learning approach that utilizes attention-based LSTM for stock market prediction using minimal data.
    \item \textbf{STLAT} \cite{li2022Slat}: Attention-based LSTM that utilizes adversarial training, data selection and transfer learning. Current state-of-the-art. 
    \item \textbf{STLAT-AT} \cite{li2022Slat}: STLAT model without the use of Adversarial Training.
    \item \textbf{STLAT-TL} \cite{li2022Slat}: STLAT model without the use of Transfer Learning.
    \item \textbf{STST-MLP-T}: Our base transformer-encoder model with traditional embeddings and MLP classifier. 
    \item \textbf{STST-T}: Our base transformer-encoder model without spatiotemporal embeddings (temporal only).
    \item \textbf{STST-MLP}: Our transformer-encoder model without the LSTM-MLP classifier.
    \item \textbf{STST}: Our transformer-encoder model utilizing both spatiotemporal embeddings and LSTM/MLP classifier.
\end{itemize}

\newcommand\scalemath[2]{\scalebox{#1}{\mbox{\ensuremath{\displaystyle #2}}}}

Following previous work on stock market prediction, we utilize two different metrics for evaluating prediction performance. The first metric is accuracy, which tells us overall how many test samples the model predicted correctly. In addition, we also use the Matthew's Correlation coefficient (MCC), a metric very commonly used for comparing methods for stock movement prediction. The MCC metric calculates the Pearson product-moment correlation coefficient between actual and predicted values. MCC helps give a more accurate result when classes are not balanced. For both metrics, a higher score indicates better model performance. The formulas for accuracy and MCC can be seen below in equations 2 and 3.

\begin{equation}
\scalemath {0.8}{Accuracy = \frac{TP + TN}{TP + TN + FP + FN}}
\end{equation}

\begin{equation}
\scalemath {0.8}{MCC = \frac{TP 
\times TN  - FN \times FP}{\sqrt{(TP + FP) \times (TP + FN) \times (TN + FP) \times (TN + FN)}}}
\end{equation}

\subsection{Q2: Ablation Study}

On top of the STST model that we outline above, we perform an ablation study to outline the importance and effectiveness of the LSTM and spatiotemporal embedding layers for our overall model performance. As a baseline, the first model that we train utilizes traditional temporal attention and an MLP classifier. This model is referred to as STST-MLP-T. Next, we train a variant using an LSTM-MLP classifier and traditional temporal attention. The purpose of this model is to show the effectiveness of the LSTM-MLP classifier on the overall results for STST. This model is referred to going forward as STST-T. Finally, the last variant that we test includes the spatiotemporal embedding layer paired with only an MLP classifier. This model is used to show the importance of spatiotemporal attention in the STST model. This model is referred to as the STST-MLP. These results are compared against the overall STST approach to outline how the use of both the spatiotemporal embedding and LSTM-MLP classifier are crucial for the model's overall performance. 

\def\arraystretch{1.2}
\begin{table*}[!htb]
\centering
\begin{tabular}{llllcclllcc}
\hline \hline
\textbf{}                 &  &  &  & \multicolumn{1}{l}{\textbf{ACL18}} & \multicolumn{1}{l}{\textbf{}} &  &                      &  & \multicolumn{1}{l}{\textbf{KDD17}} & \multicolumn{1}{l}{} \\ \cline{5-6} \cline{10-11} 
\multicolumn{1}{c}{Model}           &  &  &  & Accuracy & MCC    &  & \multicolumn{1}{c}{} &  & Accuracy & MCC \\ \hline \hline
Random                              &  &  &  & 50.102   & 0.002  &  &  &  & 49.850 & -0.004 \\
LSTM                                &  &  &  & 53.876   & 0.101  &  &  &  & 52.679 & 0.027  \\ \hline
Adv-ALSTM \cite{feng2018advALSTM}   &  &  &  & 57.200    & 0.148  &  &  &  & 53.050 & 0.052  \\
DTML \cite{yoo2021LstmTransformer}  &  &  &  & 57.440    & 0.191  &  &  &  & 53.530 & 0.073  \\ 
MAN-SF \cite{sawhney2020mansf}      &  &  &  & 60.800   & 0.195  &  &  &  & 54.150 & 0.062  \\ 
IDTLA \cite{he2022InstanceBased}    &  &  &  & 58.200   & 0.168  &  &  &  &  ---   &  ---   \\ 
STLAT-AT \cite{li2022Slat}          &  &  &  & 60.610	& 0.224  &  &  &  & 55.730 & 0.1110 \\ 
STLAT-TL \cite{li2022Slat}          &  &  &  & 58.780	& 0.170  &  &  &  & 54.570 & 0.0864 \\  
STLAT \cite{li2022Slat}             &  &  &  & \textbf{64.560}   & \textbf{0.297}  &  &  &  & \textbf{58.270} & \textbf{0.159}  \\ \hline
STST-MLP-T                          &  &  &  & 56.624   & 0.126  &  &  &  & 52.860 & 0.045  \\ 
STST-T                              &  &  &  & 59.478   & 0.186  &  &  &  & 53.357 & 0.057  \\        
STST-MLP                            &  &  &  & 62.509   & 0.249  &  &  &  & 54.319 & 0.075  \\
STST                                &  &  &  & 63.707   & 0.268  &  &  &  & 56.879 & 0.128  \\ \hline \hline           
\end{tabular}
\caption{STST results compared to other baseline and state-of-the-art approaches. \label{model-comparison}}
\end{table*}

\subsection{Q3: Investment Simulation}

For the final experiment, we utilize the STST model in a simulated market trading scenario to determine how effective its predictions are for use while trading. To do this, we create two simulated environments using the test data from the ACL18 and KDD17 datasets. The simulated agent starts with an initial cash amount of \$10000 USD. At the end of each day, the agent balances its stock portfolio evenly based on the top five highest-scoring positive predictions for the next day. If fewer than five positives exist, then it will evenly distribute the predictions among the existing positives. Additionally, if no stocks are predicted to go up, then the agent will hold off and wait for the next day. At the end of the test period, the cumulative and annualized returns are gathered. The results are then compared against the performance of the S\&P 500 index for both datasets to evaluate the simulated agent's overall performance. 

\section{Results}

We performed a grid-based search for the STST model over a large set of potential hyper-parameters, as seen in Table \ref{tbl:paramsearch}. The hyper-parameters that we found to give the best results can be seen in Table \ref{hyper-results}. The majority of the parameters stayed the same between the two datasets, with the exception of a few. The first parameter that differed was the learning rate. We found that the model performed better on the KDD dataset with a smaller learning rate of $2.5e^{-6}$. The next difference was the number of encoders. The best results for STST against the ACL18 dataset were found using four encoder layers, while the best results for the KDD17 dataset were found with eight encoders. Finally, the last difference that we found between the two during our training was the number of LSTM layers. We employed two stacked LSTM layers for ACL18 as opposed to three for KDD17. The reason for these differences is likely due to the size of the datasets. The KDD17 dataset spans a significantly longer range of time. This poses a much more difficult modeling task for machine learning models since the market evolves significantly over time, making it much harder to generalize.

\def\arraystretch{1.0}
\setlength{\textfloatsep}{0.5cm}
\begin{table}[!htp]
\centering
\begin{tabular}{l||lcllc}
\hline \hline
\multicolumn{1}{c}{Parameter} & & \textbf{ACL18} &  &  & \textbf{KDD17} \\ \hline \hline
learning\_rate                & &  $5.35e^{-6}$  & & &   $2.5e^{-6}$ \\
batch\_size                   & &  32       & & &   32     \\
warmup\_steps                 & &  25000    & & &   25000  \\
context\_window\_size         & &  32       & & &   32     \\ 
n\_encoders                   & &  4        & & &   8      \\
n\_heads                      & &  4        & & &   4      \\
d\_model                      & &  64       & & &   64     \\
d\_ff                         & &  2048     & & &   2048   \\
ff\_dropout                   & &  0.3      & & &   0.3    \\
attn\_dropout                 & &  0.1      & & &   0.1    \\
n\_lstm\_layers               & &  2        & & &   3      \\
d\_lstm\_hidden               & &  256      & & &   256    \\
norm\_type                    & &  power    & & &   power  \\
norm\_placement               & &  post     & & &   post   \\ \hline \hline
\end{tabular}
\caption{Hyper-parameter selection for the STST model. \label{hyper-results}}
\end{table}

\begin{figure*}[!htp]
\begin{tabular}{ll}
\includegraphics[trim={0.5cm 0cm 0 0cm},clip,scale=0.32]{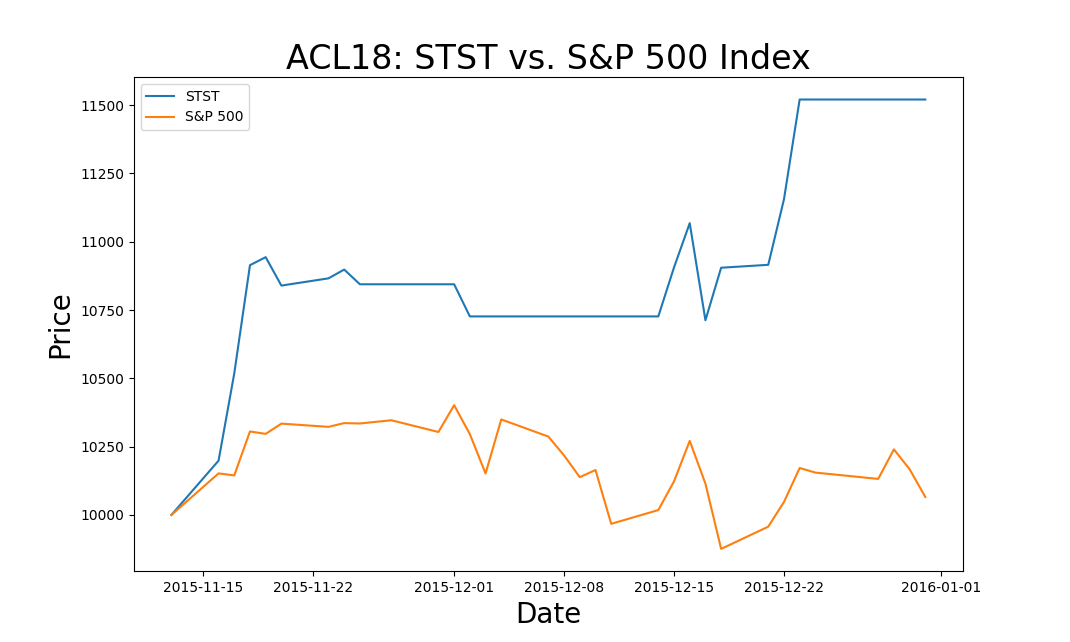}
&
\includegraphics[trim={0.5cm 0 0 0.2cm},clip,scale=0.32]{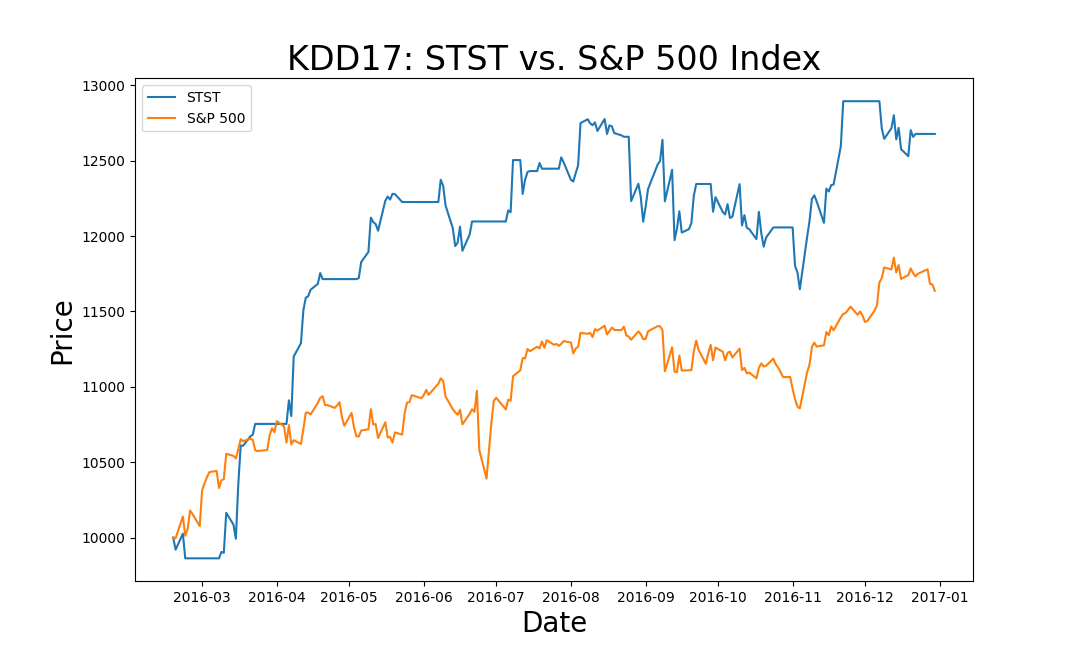}
\end{tabular}
\caption{Investment Simulation cumulative profits against compared with the S\&P 500 Index for the the ACL18 (left) and KDD17 datasets (right)}
\label{fig:simulation-results}
\end{figure*}

\subsection{Model Performance}

The STST approach performed well against both datasets, with an overall accuracy and MCC score of 63.707\% and 0.268 for the ACL18 dataset and 56.879\% and 0.128 for the KDD17 dataset, while utilizing spatiotemporal embeddings and an LSTM classifier. The comparison of STST with other baseline models can be seen in Table \ref{model-comparison}. Our model surpassed the results obtained by all but one of the previous state-of-the-art approaches without the need for external features or the use of correlating multiple assets. Unfortunately, its overall performance was slightly worse than the full STLAT model \cite{li2022Slat}, which obtained an accuracy and MCC score of 64.560\% and 0.297 for the ACL18 dataset and 58.270\% and 0.159 for the KDD17 dataset. While not performing better than the full STLAT approach, our STST approach was able to significantly out-perform both the STLAT variant without transfer learning (STLAT-TL) with accuracy improvements of 5.11\% and 6.84\% datasets, as well as the variant that does not utilize adversarial training (STLAT-AD) with accuracy improvements of 2.10\% and 4.26\% for the ACL18 and KDD17 datasets, respectively.

\subsection{Ablation Study}

For the ablation study, we tested three variants of the STST model to show both the necessity and effectiveness of the spatiotemporal attention and LSTM classifier in our resultant model. The first variant we tested served as a baseline, utilizing a transformer with traditional attention and an MLP classifier. It obtained accuracies of 56.624\% and 52.860\% against the ACL18 and KDD17 datasets, respectively. The next variant we tested was the STST with only temporal attention and an LSTM-MLP classifier (STST-T). The addition of the LSTM allowed this variant to perform significantly better than the baseline, with accuracies of 59.478\% and 53.357\% against each dataset, up to a 5.0\% increase over the baseline STST model. Finally, we tested the variant with spatiotemporal embeddings and an MLP classifier (STST-MLP). This model achieved accuracies of 62.509\% and 54.319\% for each dataset, an increase of up to 10.4\% over the baseline model. For the ACL18 and KDD17 datasets, the overall model with both spatiotemporal embeddings and an LSTM-MLP classifier was able to improve performance over the baseline by 12.51\% and 6.64\%, respectively. 

\subsection{Investment Simulation}

To test the model's overall real-life applicability for use in trading, we performed investment simulations. The STST model is utilized by selecting the model's top-5 positive predicted assets at each timestep. This is used to rebalance a portfolio of stocks in order to maximize profits. The graph of the cumulative profits of STST compared to the S\&P 500 index can be seen in Figure \ref{fig:simulation-results}. For the ACL18 test ranges, STST had a return of 15.21\% while the S\&P 500 index had a return of 0.66\%. Due to the short length of time in the ACL18 test dataset, this amounts to an annualized return of 199\%. With the KDD17 dataset, STST had a return of 26.78\% while the S\&P 500 index had a return of 16.36\%, resulting in an annualized return of 31.24\%. Between the two test datasets, our STST model was able to obtain a minimum of 10.41\% higher returns than the S\&P 500 index, one of the most common benchmarks used for evaluating trading strategies. Analyzing the charts for each dataset, it is easy to see that in most of the places where the STST model strategy chose not to trade, meaning that it did not predict any positive moving stocks, holding onto the S\&P 500 index lost money. This shows that the STST model is able to adequately predict areas where the stock price is decreasing. 

\section{Discussion}

Overall, our model performed very well in comparison to the previous and current state-of-the-art approaches. STST was able to outperform all previous state-of-the-art approaches and came within 0.853\% and 1.391\% of the current state-of-the-art approach STLAT for the ACL18 and KDD17 datasets, respectively. 

It is important to note that the authors mention that the focus of the STLAT approach is on the training process and not model selection. Consequently, it focuses mostly on the process of selective sampling for fine-tuning and the use of adversarial training as opposed to determining the optimal base model. In their research, the base model of STLAT is the same as the Adv-ALSTM, which our model performed significantly better than. Li et al. provided the results of their approach without the use of transfer learning (STLAT-TL) as well as adversarial training (STLAT-AT), of which our model was able to outperform both by a minimum of 3.097\% and 1.149\% against ACL18 and KDD17, respectively. In addition, our model outperforms the previous state-of-the-art approaches, including Adv-ALSTM, DTML and MAN-SF, each of which is more focused on model selection. Therefore, while the STLAT training approach may yield better results overall, STST provides a better basis for stock movement prediction. Unfortunately, the STLAT approach is not publicly available, therefore, we are unable to perform additional comparisons with their approach.

Following the comparison of the baseline models, we performed an ablation study to determine the overall effectiveness of the spatiotemporal embedding and LSTM classifier towards our model's overall performance. Based on our results, we found that the spatiotemporal embedding provided up to 10.4\% improvement and the LSTM classifier provided improvements up to 5.0\% over the baseline STST model. From this, we can conclude that the spatiotemporal embedding provides a much larger overall benefit for the STST approach. However, while the performance increase for the spatiotemporal embedding is greater, we cannot discount the need for the LSTM-MLP classifier. Combined, both models provide a staggering performance improvement of 12.5\% over the baseline. Therefore, we can conclude that both the spatiotemporal embedding and the LSTM classifier are necessary for the STST approach to obtain the best results. 

One important aspect to discuss is the disparity between the accuracies obtained for the ACL18 and KDD17 datasets, which can be observed in the accuracy and MCC scores obtained by both our model and the other baseline approaches seen in Table \ref{model-comparison}. This is most likely due to the length of time contained within each dataset. The KDD17 dataset spans 10 years, from 2007 to 2017, whereas the ACL18 dataset only spans two years, from 2014 to 2016. Over history, the stock market has been constantly evolving due to changes in macroeconomics, foreign policy, technological advances, market agents and much more. When you observe a wider time frame, you start to see how the market changes and how particular patterns or evaluation techniques may no longer be as useful as they previously were \cite{plastun2021evolution}. This is most likely the culprit causing the performance difference between the two datasets. For the ACL18 dataset, the date range is small and likely does not have a large enough window into the history of the stock market to observe different types of market behavior. In addition, the ACL18 dataset only contains stocks between 2014 and 2016, which was a primarily bullish time period without a lot of turmoil. For the KDD17 dataset, it spans a longer period of time starting in 2007. Therefore, the dataset captures the ``Great Recession'' of 2008-2009, which had a drastic influence on the market for years to come. Therefore, the KDD17 dataset is significantly more likely to contain many different periods of time that exhibit different types of behavior. It is likely substantially more difficult for a machine learning model to generalize over such a diverse sampling.

\section{Future}
There are a number of different areas that we plan to research in the future in order to further improve the performance of the STST model. The first area for future research is the use of adversarial training. Numerous previous state-of-the-art approaches have found great success while using adversarial training \cite{feng2018advALSTM} \cite{li2022Slat} for stock movement prediction. The use of adversarial training through added random perturbations can be used to help improve the model's ability to generalize and overcome the noisy stochastic nature of the stock market. 

Another area that we want to further expand research on is the ability to leverage correlated assets and stock indices to give a broader view of the market, which will allow for better informed price predictions. While all stocks have their own set of factors that can impact their price, there are often many macroeconomic factors or political happenings that can impact multiple stocks across a sector or the entire market. However, some stocks may be more reactive to these types of changes due to a number of factors, and some price changes may lag behind others. It is possible to leverage this in order to make predictions as to the direction a stock will move. Trading around this notion is often referred to as ``arbitrage.'' We could utilize this aspect using STST by using the spatio-temporal transformer to encode multiple different stocks within a common sector. The output for each stock can then be concatenated together, creating a multi-context view of the market in a manner similar to DTML's approach \cite{yoo2021LstmTransformer}. This can then be fed through an LSTM or another transformer to correlate patterns and make a final prediction. 

\section{Conclusion}

With the potential for making large profits, a vast amount of research has been focused on trying to predict various aspects of the stock market, including price movement and trading signals. While many traders attempt this manually, the task becomes quickly unmanageable for a human to perform accurately due to the amount of data that should be considered when making predictions. These difficulties, along with the rise of machine learning, have led to a massive influx of research aimed at applying machine learning within the stock market realm. Unfortunately, the relationships that are contained in stock data are complex due to the stochastic nature of the market and are often very difficult to extract. To combat this, we propose STST, a spatiotemporal transformer-encoder paired with an LSTM for extracting complex relationships from stock data. We showcase its ability on the task of stock price movement prediction. STST beats the previous four state-of-the-art approaches but came up just shy of the current state-of-the-art by 0.853\% and 1.391\% for the ACL18 and KDD17 datasets, respectively. While unable to beat STLAT performance, we demonstrate that our base model has a greater capacity for learning than previously used models for stock market prediction. STST provides a strong basis for future research, with numerous potential prospects that could easily allow the model to overcome the current state-of-the-art. On top of the model's performance, we show the real-life applicability of our model in the stock market through simulation, with the ability to obtain returns significantly higher than the S\&P 500 market index.

\balance

\bibliography{references.bib}
\bibliographystyle{aaai}
\end{document}